\documentclass{article}
\usepackage{arxiv}
\usepackage[utf8]{inputenc} 
\usepackage[T1]{fontenc}    
\usepackage{hyperref}       
\usepackage{url}            
\usepackage{enumitem}
\setlist[itemize]{leftmargin=.5cm}
\setlist[enumerate]{leftmargin=.5cm}
\setlist[description]{leftmargin=.5cm, labelindent=\parindent}

\usepackage{xcolor}
\usepackage{graphicx}
\usepackage[capposition=top]{floatrow}
\definecolor{codegreen}{rgb}{0,0.6,0}
\definecolor{codegray}{rgb}{0.5,0.5,0.5}
\definecolor{codepurple}{rgb}{0.58,0,0.82}
\definecolor{backcolour}{rgb}{0.95,0.95,0.92}

\usepackage{listings}
\lstdefinestyle{mystyle}{
    backgroundcolor=\color{backcolour},
    commentstyle=\color{codegreen},
    keywordstyle=\color{magenta},
    numberstyle=\tiny\color{codegray},
    stringstyle=\color{codepurple},
    basicstyle=\ttfamily\footnotesize,
    breakatwhitespace=false,
    breaklines=true,
    captionpos=b,
    keepspaces=true,
    numbers=left,
    numbersep=5pt,
    showspaces=false,
    showstringspaces=false,
    showtabs=false,
    tabsize=2
}
\usepackage{booktabs}       
\usepackage{amsfonts}       
\usepackage{nicefrac}       
\usepackage{microtype}      
\usepackage{adjustbox}

\usepackage{tikz}
\usetikzlibrary{arrows,shapes.geometric,chains,matrix,positioning,scopes,decorations.pathreplacing, angles,quotes}

\newcommand\litem[1]{\item{\bfseries #1}} 

\title{sktime: A Unified Interface for Machine Learning with Time Series}

\author{
    Markus Löning \\
    The Alan Turing Institute \\
    \And
    Anthony Bagnall \\
    University of East Anglia \\
    \And
    Sajaysurya Ganesh \\
    University College London \\
    \And
    Viktor Kazakov \\
    University College London \\
    \And
    Jason Lines \\
    University of East Anglia \\
    \And
    Franz J.~Kir\'{a}ly\thanks{Corresponding author: \texttt{fkiraly@turing.ac.uk}} \\
    The Alan Turing Institute\\
}

\begin{document}
\maketitle

\begin{abstract}
We present sktime -- a new scikit-learn compatible Python library with a unified interface for machine learning with time series. Time series data gives rise to various distinct but closely related learning tasks, such as forecasting and time series classification, many of which can be solved by reducing them to related simpler tasks. We discuss the main rationale for creating a unified interface, including reduction, as well as the design of sktime's core API, supported by a clear overview of common time series tasks and reduction approaches.
\end{abstract}

\section{Introduction}
Data scientific tasks beyond the standard tabular setting are one of the major challenges of contemporary machine learning. sktime\footnote{\url{https://github.com/alan-turing-institute/sktime} 
}
is a new open-source Python library for machine learning with time series. Our goal is to extend existing machine learning capabilities, most notably scikit-learn \cite{Buitinck2013}, to the temporal data setting by providing a unified interface for several time series learning tasks.

Time series data is ubiquitous in many applications. Examples include sensor readings from industrial processes, spectroscopy wave length data from chemical samples, or bed-side monitor medical data from patients. There is a broad variety of distinct but closely related learning tasks that arise in such contexts, including time series classification, forecasting and annotation among others. In section \ref{sec:learning}, we give a more detailed overview of time series tasks. The ambition of the project is to design and implement an API (application programming interface) that unifies these tasks.

A plethora of time series toolboxes exists that provide rich interfaces to specific model classes (ARIMA/filters \cite{Perktold2010, taylor2016pyflux}, neural networks \cite{alexandrov2019gluonts}), or framework interfaces to isolated time series modelling tasks (forecasting \cite{taylor2018forecasting, Guecioueur2018}, feature extraction \cite{Kanter2015, tavenard2017tslearn, Christ2018}, annotation \cite{burns2018seglearn}, classification \cite{burns2018seglearn, tavenard2017tslearn}).\footnote{For a more extensive and regularly updated overview of Python time series related libraries, see \url{https://github.com/alan-turing-institute/sktime/wiki/Related-software}.} Nevertheless, open-source machine learning capabilities for time series are still limited and existing libraries are often incompatible with each other. To the best of our knowledge, we are the first to present a unified interface that can explicitly represent and link multiple distinct tasks. 

The main rationale for creating a unified API, as opposed to separate task-specific interfaces, is as follows:
First, many time series learners are highly composite and often involve reduction from complex learning task (e.g.\ time series segmentation) to related simpler tasks (e.g.\ supervised learning). We describe exemplar reduction approaches with time series data in greater depth in section \ref{sec:reduction}. A unified and composable interface for different tasks enables us to encapsulate reduction approaches as meta-estimators, exposing their implicit modeling choices as tunable hyper-parameters, and thus enabling us to easily evaluate and compare them against other strategies.
Second, the current lack of a unified API leads to unnecessary code replications and often error prone and statistically inappropriate reductions to those tasks that existing off-the-shelf toolboxes can deal with (e.g. reductions to tabular data supported by scikit-learn).
A single API reduces confusion and enables us to focus on providing advanced time series analysis capabilities for researchers and practitioners. In addition, many tasks require common functionality such as distance measures and preprocessing routines. Providing them in a consistent and modular interface allows us to re-utilise them across different settings.

As of now, sktime includes state-of-the-art algorithms for time series classification, additional modular functionality for reduction, pipelining, ensembling and data transformations, as well as forecasting methods and benchmarking tools.

We first present in section \ref{sec:learning} an overview of the most common time series tasks, and then discuss, in section \ref{sec:reduction}, reduction as an approach to solving these tasks. Section \ref{sec:interface} outlines the key design features built into the core interface. Section \ref{sec:features} describes the currently available functionality. We conclude by previewing future work in section \ref{sec:conclusion}.

\section{Taxonomy of time series learning tasks}
\label{sec:learning}
sktime provides tooling for learning with time series, i.e.\ data observed at a finite number of known time points. More precisely, whenever we refer to time series, we consider them explicitly consisting of both (i) time points at which they are observed and (ii) observations at those time points. In ad-hoc short-hand notation, we write, for example, $x(t_1), x(t_2), ..., x(t_T)$ for observations at time points $t_1, t_2, ..., t_T$, and $\textbf{x}$ for the time series object that contains exactly that information.\footnote{Formally, our data model for time series is that of a list of time/value pairs, or equivalently, a time indexed list or dictionary, with a finite set of indices.} An intrinsic characteristic of time series is that observations within time series are statistically dependent in assumed generative processes (which we avoid to introduce here to keep notation simple). Due to this dependency, time series data does not naturally fit into the standard machine learning framework for tabular (or cross-sectional) data, which implicitly assumes observations to be independent and identically distributed (i.i.d.). Consequently, many toolboxes for learning on tabular data, such as scikit-learn, consider learning with time series out of scope \cite{Buitinck2013}.

When learning with time series, it is important to understand the different forms such data may take. The data can come in the form of a single (or univariate) time series; but in many applications, multiple time series are observed. It is crucial to distinguish the two fundamentally different ways in which this may happen:
\begin{itemize}
    \litem{Multivariate time series data}, where two or more variables are observed over time, with variables representing \textit{different kinds of measurements} within a single experimental unit;
    \litem{Panel data}, sometimes also called longitudinal data, where multiple \textit{independent instances} of the \textit{same kinds of measurements} are observed, e.g.\ time series from multiple industrial processes, chemical samples or patients.\footnote{One complication is that observed time points may vary across variables and/or instances. While time-heterogeneous settings are not covered by our notation, they are covered by the sktime interface.}
\end{itemize}
In multivariate data, it is implausible to assume the different univariate component time series are i.i.d. In panel data, the i.i.d.\ assumption applied to the different instances is plausible, while time series observations within a given instances may still depend on adjacent observations. In addition, panel data may be multivariate, which corresponds to i.i.d.\ instances of multivariate time series. In this case, the different instances are i.i.d., but the univariate component series within an instance are not. This richness of generative scenarios is mirrored in a richness of learning tasks applicable to such data. We later show that these tasks are closely related through reduction, but first highlight some of the most common ones here:
\begin{itemize}
    \litem{Time series regression/classification.} We observe $N$ i.i.d.\ panel data training instances of feature-label pairs $(\textbf{x}_i, y_i)$, $i=1\dots N$.
    Each instance of features is a time series $\textbf{x}_i = (x_i(t_1) \dots x_i(t_T))$. The task is to use the training data to learn a predictor $\hat{f}$ that can accurately predict a new target value, such that $\hat{y} = \hat{f}(\textbf{x}_{*})$ for a new input time series $\textbf{x}_{*}$. For regression, $y_i\in\mathbb{R}$. For classification, $y_i$ takes a value from a finite set of class values. Additionally, time-invariant features may be present. Compared to the tabular supervised setting, the only difference is that some features are time series, instead of being only primitives (e.g.\ numbers, categories or strings). Important sub-cases are (i) equally spaced observation times and (ii) equal length time series. For an overview of time series classification, see  \cite{bagnall17bakeoff, fawaz18deep}.
    \litem{Classical forecasting.} Given past observations $\textbf{y} = (y(t_1)\dots y(t_T))$ of a single time series, the task is to learn a forecaster $\hat{f}$ which can make accurate temporal forward predictions $\hat{y} = \hat{f}(h_j)$ of observations at given time points $h_1\dots h_H$ of the forecasting horizon, where $\hat{\textbf{y}} = (\hat{y}(h_1) \dots \hat{y}(h_H))$ denotes the forecasted series. No i.i.d.\ assumption is made. Variants may be distinguished by the following:
    (i) whether one observes additional related time series (multivariate data);
    (ii) for multivariate data, whether one forecasts a single series or multiple series jointly (exogeneity vs vector forecasting) \cite{Lutkepohl2005};
    (iii) whether the forecasting horizon lies in the observed time horizon (in-sample predictions), in the future of the observed time series (forecasting), or for multivariate data, only in the future of the target variable but not the exogenous variables (nowcasting);
    (iv) whether there is a single time point to forecast ($H=1$) or not (single-step vs multi-step); (v) whether the forecasting horizon is already known during training or only during forecasting (functional vs discrete forecast). For an overview of classical forecasting, see \cite{Box2015, Brockwell2016, hyndman2018forecasting, Hyndman25years}.
    \litem{Supervised/panel forecasting.} We observe $N$ i.i.d.\ panel data training instances $(\textbf{y}_i)$, $i=1 \dots N$. Each instance is a sequence of past observations $\textbf{y}_i = (y_i(t_1)\dots y_i(t_T))$. The task is to use the training data to learn a supervised forecaster $\hat{f}$ that can make accurate temporal forward predictions $\hat{y}_i = \hat{f}(\textbf{y}_{*}, h_j)$ for a new instance $ \textbf{y}_{*}$ at given time points $h_1\dots h_H$ of the forecasting horizon, where $\hat{y}_i = (\hat{y}_i(h_1) \dots \hat{y}_i(h_H))$ is the forecasted series.
    Variants include panel data with additional time-constant features and the same variants as found in classical forecasting. For an overview, see \cite{Baltagi2008, Wooldridge2010, Diggle2013a}.
    \litem{Time series annotation.} For given observations $\textbf{x} = (x(t_1)\dots x(t_T))$ of a single time series, the task is to learn an annotator that accurately predicts a series of annotations $\hat{\textbf{y}} = (\hat{y}(a_1)\dots \hat{y}(a_A))$ for the observed series $\textbf{x}$, where $a_1\dots a_A$ denotes the time indices of the annotations. The task varies by value domain and interpretation of the annotations $\hat{\textbf{y}}$ in relation to $\textbf{x}$: (i) in change-point detection, $\hat{\textbf{y}}$ contains change points and the type of change point \cite{guralnik1999event}; (ii) in anomaly detection, the $a_j$ are a sub-set of the $t_j$ and indicate anomalies, possibly with the anomaly type \cite{Box2015}; (iii) in segmentation, the $a_j$ are interpreted to subdivide the series $\textbf{x}$ into segments, annotated by the type of segment \cite{keogh2004segmenting}. Time series annotation is also found in supervised form, with partial annotations within a single time series, or multiple annotated i.i.d.\ panel data training instances \cite{Dietterich2002, graves2012supervised}. 
\end{itemize}


\section{Reductions with time series}
\label{sec:reduction}
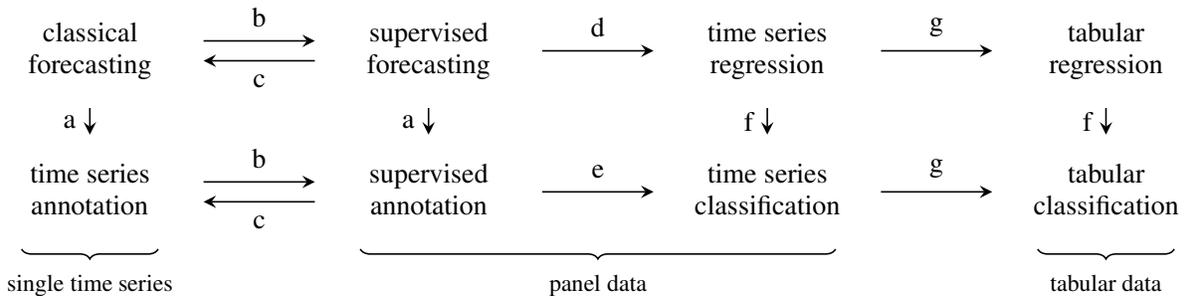
\begin{figure}[!htb]
\centering
\caption{Stylised overview of time series reduction approaches}
\label{fig:reductions}
    \begin{adjustbox}{width=\textwidth}
    \begin{tikzpicture}

    \tikzstyle{task} = [rectangle, minimum width=2cm, minimum height=1cm, text centered]
    \tikzstyle{arrow} = [->,>=stealth]
    \tikzstyle{every node}=[font=\scriptsize]

    \node (classical_forecasting) [task] {
        \begin{tabular}{c}
            classical \\
            forecasting
        \end{tabular}
    };
    \node (supervised_forecasting) [task, right of=classical_forecasting, xshift=2cm] {
        \begin{tabular}{c}
            supervised \\
            forecasting
        \end{tabular}
    };
    \node (time_regression) [task, right of=supervised_forecasting, xshift=2cm] {
        \begin{tabular}{c}
            time series \\
            regression
        \end{tabular}
    };
    \node (tabular_regression) [task, right of=time_regression, xshift=2cm] {
        \begin{tabular}{c}
            tabular \\
            regression
        \end{tabular}
    };
    \node (annotation) [task, below of=classical_forecasting, yshift=-.25cm] {
        \begin{tabular}{c}
            time series \\
            annotation
        \end{tabular}
    };
    \node (supervised_annotation) [task, right of=annotation, xshift=2cm] {
        \begin{tabular}{c}
            supervised \\
            annotation
        \end{tabular}
    };
    \node (time_classification) [task, right of=supervised_annotation, xshift=2cm] {
        \begin{tabular}{c}
            time series \\
            classification
        \end{tabular}
    };
    \node (tabular_classification) [task, right of=time_classification, xshift=2cm] {
        \begin{tabular}{c}
            tabular \\
            classification
        \end{tabular}
    };

    \draw [arrow] (classical_forecasting.5) -- node[anchor=south] {b} (supervised_forecasting.175);
    \draw [arrow] (supervised_forecasting.185) -- node[anchor=north] {c} (classical_forecasting.355);
    \draw [arrow] (supervised_forecasting) -- node[anchor=south] {d} (time_regression);
    \draw [arrow] (time_regression) -- node[anchor=south] {g} (tabular_regression);

    \draw [arrow] (annotation.5) -- node[anchor=south] {b} (supervised_annotation.175);
    \draw [arrow] (supervised_annotation.185) -- node[anchor=north] {c} (annotation.355);
    \draw [arrow] (supervised_annotation) -- node[anchor=south] {e} (time_classification);
    \draw [arrow] (time_classification) -- node[anchor=south] {g} (tabular_classification);

    \draw [arrow] (classical_forecasting) -- node[anchor=east] {a} (annotation);
    \draw [arrow] (supervised_forecasting) -- node[anchor=east] {a} (supervised_annotation);
    \draw [arrow] (time_regression) -- node[anchor=east] {f} (time_classification);
    \draw [arrow] (tabular_regression) -- node[anchor=east] {f} (tabular_classification);

    \draw[decoration={brace,mirror}, decorate] (annotation.220) -- node[below=.1cm] {\tiny single time series} (annotation.320);
    \draw[decoration={brace,mirror}, decorate] (supervised_annotation.220) -- node[below=.1cm] {\tiny panel data} (time_classification.320);
    \draw[decoration={brace,mirror}, decorate] (tabular_classification.220) -- node[below=.1cm] {\tiny tabular data} (tabular_classification.320);

    \end{tikzpicture}
    \end{adjustbox}
\vspace{-0.4cm}
\floatfoot{\textit{Notes}: (a) annotate time series with future values, (b) rolling window method to convert single series into panel data with multiple output time periods \cite{Bontempi2012}, (c) ignore training set (e.g.\ fit forecaster on test set only) or use training set for model selection, (d) iterate over output periods, optionally time binning/aggregation of output periods \cite{Bontempi2012}, (e) rolling window method to convert single series into panel data with single output period \cite{Dietterich2002}, (f) discretise output into one or more bins, (g) feature extraction \cite{Fulcher2017, Christ2018} or time binning/aggregation of input time points.}
\end{figure}

While these tasks define distinct learning settings, they are closely related, which enables us to solve them via reduction. Reduction essentially decomposes a given task into simpler tasks so that solutions to the simpler tasks can be composed to give a solution to the original task. A classical example of reduction in tabular supervised learning is one-vs-all classification, reducing $k$-way multi-category classification to $k$ binary classification tasks \cite{beygelzimer2008machine, beygelzimer2005weighted, beygelzimer2015learning}. For time series, a common example of reduction is to solve classical forecasting through time series regression via a sliding window approach and iteration over the forecasting horizon \cite{Bontempi2012}. 
Many reduction approaches are possible with time series, we highlight some of the most important ones in figure \ref{fig:reductions}.

Reduction offers several key advantages with regard to API design \cite{beygelzimer2005error, beygelzimer2008machine}: First, reductions convert any algorithm for a particular task into a learning algorithm for the new task. Any progress on the base algorithm immediately transfers to the new task, saving both research and software development effort. Second, reductions are modular and composable. Applying some reduction approach to $n$ base algorithms gives $n$ new algorithms for the new task. Reductions can be composed to solve more complicated problems, e.g.\ first reducing forecasting to time series regression which in turn can be reduced to tabular regression. Finally, reductions also help us better understand the relationship between tasks and reduce confusion between them.


\section{API design}
\label{sec:interface}


The main goal of sktime is to create a unified API for multiple time series tasks, extending the common scikit-learn interface to the temporal setting, while staying close to its syntax and logic whenever possible. Following scikit-learn's API allows us to re-utilise many of the algorithms available in scikit-learn, which is especially useful because of reduction to tabular tasks and because many specialised algorithms for time series are composites with tabular supervised learning algorithms as their components. The key design features of the sktime API are as follows:

\subsection{Data representation}
Any machine learning library relies fundamentally on some data representation and the lack of powerful data structures for time series data has arguably been one of the main reasons for the lack of a unified interface. In order to combine different tasks and data formats, sktime requires a data container capable of handling multivariate and panel data with additional time-constant features, including time-heterogeneous time series, where the observed lengths and time points varies across instances and/or variables. While pandas \cite{McKinney2011} handles time series data, it is intended to store time series only in the long format, with rows representing time points and columns representing variables, precluding time-heterogeneous data. Technically, however, it is possible to store arbitrary types in the cells of pandas containers. Inspired by xpandas \cite{davydov2018xpandas}, we chose to exploit this feature and represent time series data in a nested format, with rows representing i.i.d.\ instances and columns representing different variables as before, but with cells now no longer representing only primitives but also entire time series. The reason for this choice is twofold: First, we can still make use of pandas, one of the most comprehensive and efficient data container in Python, while at the same time having a consistent data representation across different tasks which is flexible enough to handle multivariate, panel and time-heterogeneous data. Second, as rows still represent i.i.d.\ instances, this representation allows us to reuse many of the existing functionality in scikit-learn. Alternatives we considered but ultimately set aside include three-dimensional NumPy arrays \cite{VanderWalt2011} as used in \cite{tavenard2017tslearn} and xarray \cite{Hoyer2017, xarray_v0_8_0}, an extension of pandas to panel data, both however only support time-homogeneous data.\footnote{Other pandas-based data containers we considered include entity sets from Featuretools \cite{Kanter2015} and pysf \cite{Guecioueur2018}.}

\subsection{Task-specific estimators}
We follow scikit-learn \cite{Varoquaux2015, Pedregosa2001} and Weka \cite{Holmes1994, Hall2009} in adopting a uniform basic API for estimators, consisting of a fit method used for learning a model from training data and a predict method used for making predictions based on the fitted model, as well as a common interface for setting and retrieving hyper-parameters. Estimators in sktime are task specific, extending scikit-learn's regressors and classifiers to their time series counterparts as well as adding new estimators such as forecasters and supervised forecasters among others, with the same fit and predict methdods, but varying function signatures and internal behaviour.


\subsection{Transformers}
Similar to estimators, transformers have a uniform API consisting of a fit and a transform method used to transform input data, and a corresponding method for the inverse transformation if available, in addition to the common hyper-parameters interface. Since many data transformations are applicable for different tasks, we develop a unified transformer interface. To reconcile the different settings, we introduce the following kinds of transformations, distinguishing between fitting over i.i.d.\ instances and fitting over time points.
\begin{itemize}
    \litem{Tabular.} scikit-learn like transformers, which operate over i.i.d.\ instances and are fitted during training (e.g.\ principal component analysis).
    \litem{Series-to-primitives.} Operates over time points, transforms time series for each instance into a primitive number (e.g.\ feature extraction). If the transformer is fittable, it is fitted separately for each instance during both training and prediction.
    \litem{Series-to-series.} Like series-to-primitives, but output of transformation is itself a series instead of a primitive (e.g.\ Fourier transform or series of fitted auto-regressive coefficients).
    \litem{Detrending.} Operates over time points and transforms an input time series, returning a detrended time series in the same domain as the input series (e.g.\ polynomial or seasonal detrending). Detrending tranformers keep track of the time index seen in fitting, so that trends are correctly computed over new time indices when transforming new data. In forecasting, it is fitted only during training. For panel data, it can be applied like a series-to-series transformation by iterating over instances. Detrenders are designed to take forecasters as input arguments, so that they internally first fit the forecaster to the input series and then return the residuals after subtracting the in-sample forecasts from the input series. For example, to detrend a time series with an exponential smoothing forecaster in sktime, we can write:
\begin{lstlisting}[language=Python]
t = Detrender(forecaster=ExponentialSmoothingForecaster())
yt = t.fit_transform(y)
\end{lstlisting}
where the given input series \texttt{y} is transformed into \texttt{yt},   consisting of the residuals of the exponentially smoothed in-sample forecasts.
\end{itemize}

\subsection{Composition}
Many learning strategies are expressible as meta-estimators, including pipelines, ensembles, reduction approaches and model selection routines like grid-search cross-validation. As in scikit-learn, meta-estimators wrapping estimators are estimators themselves with the same uniform API, as well as support for setting and getting hyper-parameters of the wrapped estimators.

sktime is the first toolbox, as far as we know, that provides reduction strategies as composable meta-estimators with an explicit hyper-parameter interface. Reduction typically introduces implicit modelling choices (e.g.\ the window width and step length of the sliding operation when reducing forecasting to time series regression). Defining reduction approaches as meta-estimators enables us to expose these modelling choices as hyper-parameters, which allows us not only to easily compose different reduction strategies with configurable components, but also to optimise these modelling choices via common model selection techniques. For example, reducing classical forecasting to time series regression in sktime looks as follows:
\begin{lstlisting}[language=Python]
f = ReducedRegressionForecaster(regressor=RandomForestRegressor(),
                                method='recursive', window_length=2)
f.fit(y)
y_hat = f.predict(fh=fh)
\end{lstlisting}
where \texttt{y} is an input series, \texttt{fh} the forecasting horizon and \texttt{y\_hat} the forecasted series. \texttt{ReducedRegressionForecaster} is a meta-estimators with explicit, tunable hyper-parameters for determining the window length and the method for iterating over multiple output periods of the forecasting horizon (here set to the recursive approach as described in \cite{Bontempi2012}). Since \texttt{ReducedRegressionForecaster} is a forecaster itself, it could be passed into a temporal grid-search cross-validation routine \cite{bergmeir2018note} in order to optimise the hyper-parameters. Through these composition interfaces for reduction, sktime's API allows to solve a wide variety of learning tasks with a small amount of easy-to-read code.

In addition, we introduce new modular composition meta-estimators for multivariate time series. This includes column-wise ensembling, where a different estimator is applied to each time series variable inspired by scikit-learn's column transformer, and column concatenation, where multiple time series columns are concatenated into a single, long time series column.

\section{API overview}
\label{sec:features}
sktime is in an early stage of development and currently includes the following functionality:
\begin{itemize}
    \litem{Time series classification.} State-of-the-art algorithms, including
    \begin{itemize}
        \litem{Interval based.} Time series forest \cite{deng2013time} and random interval spectral ensemble~\cite{lines18hive}.
        \litem{Distance based.} Distance measures form a fundamental primitive of many time series tasks. We have implemented eight distance measures in Cython \cite{behnel2011cython} for enhanced performance and several classifiers that use them, including the Elastic Ensemble \cite{lines15elastic}, Proximity Forest \cite{lucas19proximity} and all the kernel methods described in a recent survey paper \cite{abanda19distance}.
        \litem{Shapelet based.} The shapelet package includes an implementation of the shapelet transform \cite{bostrom17binary} and prototypes for learning shapelets \cite{grabocka14learning-shapelets} and Shapelet Forest \cite{karlsson15forests}.
        \litem{Dictionary based.} The symbolic aggregate approximation (SAX)~\cite{lin07sax} and symbolic Fourier approximation (SFA) \cite{schafer12sfa} transform and the associated bag of patterns (BOP) \cite{lin12bagofpatterns} and bag of SFA symbols (BOSS) \cite{schafer15boss} classifiers.
        \litem{Deep learning.} A recent review paper described nine deep learning algorithms for time series classification \cite{fawaz18deep}. With the help of the authors, we have ported these algorithms into sktime in a deep learning extension package\footnote{\url{https://github.com/uea-machine-learning/sktime-dl}} based on keras \cite{chollet2015keras}.
    \end{itemize}
    \litem{Classical forecasting.} We have implemented a range of statistical forecasting techniques, interfacing statsmodels \cite{Perktold2010} whenever possible, and reduction strategies to utilise supervised learning algorithms.
    \litem{Transformers.} Various transformers have been implemented for segmenting time series, series-to-primitives and series-to-series feature extraction and detrending. To reduce time series regression/classification to tabular supervised learning, a transformer for time binning input series is included.
    \litem{Composition.} Pipelining for both feature and target variables following scikit-learn's API and multivariate composite strategies have been implemented.
    \litem{Benchmarking.} Inspired by the mlaut package \cite{kazakov2019machine}, sktime includes tools for automatic orchestration of prediction experiments evaluating one or more models on one or more data sets, with post-hoc statistical methods for comparing predictive performances.
\end{itemize}
The majority of the implemented algorithms has been tested for correctness against implementations in other languages and benchmarked on archive data \cite{taleoftwotoolkits2019}.
We have reproduced the results presented in a comparative benchmarking study \cite{bagnall17bakeoff} and are in the process of recreating results from forecasting benchmarking studies \cite{makridakis2019m4, makridakis2018statistical}.

\section{Conclusion and future directions}
\label{sec:conclusion}
We have discussed the main rationale for a unified interface for machine learning with time series and outlined the key features of sktime's API, including new meta-estimators for reduction and multivariate ensembling. For future development, there are several directions the sktime project aims to focus on and we are actively looking for contributors to implement task-specific interfaces and reduction approaches. At present, most methods in sktime only support data with equal length series and no missing values. We aim to extend existing functionality to cover these situations as well. In addition, the composite structure of many of the implemented time series classifiers allows us to easily refactor them into their regressor counterparts. Having designed and implemented the key building blocks of the API for time series classification/regression and classical forecasting, the next major addition to sktime will be supervised forecasting, based on a modified pysf interface \cite{Guecioueur2018}. Finally, many implemented tools in sktime (e.g.\ distance measures) can be re-utilised for related unsupervised learning task, including time series clustering and motif discovery.


\section*{Authors' contributions}
ML made key contributions to architecture and design, including composition and reduction interfaces. ML is also one of sktime's core contributors and maintainers, having implemented, and contributing to, almost all parts of it, including the overall framework, the forecasting module, and specific algorithms. ML drafted and wrote most of this manuscript, partly based on sketches and presentation content by FK.

FK conceived the project and architectural outlines, including task taxonomy, composition and reduction. FK further made key contributions to architecture and design, and contributed to writing of this manuscript.

AB implemented time series forest and the random interval spectral ensemble, and contributed design ideas and to writing of this manuscript. 

JL implemented and conceived modularised interfaces of several algorithms: distance based algorithms, including time series k-nearest-neighbours, Cython implementations of time series distance functions, and the shapelet transform.

SG contributed to the initial design and implementation of the time series classification setting, as well as implementation of the overall framework.

VK contributed to the design and implementation of the benchmarking module based on the mlaut package.



All authors reviewed the manuscript and participated in final copy-editing and proof-reading.

\section*{Acknowledgements}
We would like to thank all participants of the 2019 joint sktime/MLJ development sprint who helped implement various time series classification algorithms, including Amaia Abanda Elustondo, Aaron Bostrom, Saurabh Dasgupta, David Guijo-Rubio, James Large, Matthew Middlehurst, George Oastler, Piotr Oleśkiewicz, Mathew Smith and Jeremy Sellier. 

The first phase of development for sktime was done jointly between researchers at the University of East Anglia (UEA), University College London (UCL) and The Alan Turing Institute as part of a UK Research and Innovation (UKRI) project to develop tools for data science and artificial intelligence.

Markus Löning's contribution was supported by the Economic and Social Research Council (ESRC) [grant: ES/P000592/1], the Consumer Data Research Centre (CDRC) [ESRC grant: ES/L011840/1], and The Alan Turing Institute (EPSRC grant no. EP/N510129/1).

\small
\bibliographystyle{plain}
\bibliography{references}

\end{document}